\documentclass{article} 
\usepackage{nips12submit_e,times}
\usepackage[section]{algorithm}
\usepackage{algorithmic}
\usepackage{array}
\usepackage{amsmath}
\usepackage{amssymb}
\usepackage{amsfonts}
\usepackage{amsthm}
\usepackage{booktabs}
\usepackage{color}
\usepackage{graphicx}
\usepackage{subfigure}
\usepackage{multirow}
\usepackage{threeparttable}
\usepackage{hyperref}
\usepackage{makecell}
\usepackage{fmtcount}
\usepackage{multicol}

\numberwithin{equation}{section}

\theoremstyle{remark}

\newcommand{\cA}{\mathcal{A}}
\newcommand{\cD}{\mathcal{D}}

\newcommand{\imgX}{u}
\newcommand{\imgY}{f}
\newcommand{\imgG}{g}

\newcommand{\patchx}{x}

\newcommand{\cI}{\mathcal{I}}

\newcommand{\st}{\text{subject to}\;}

\newcommand{\R}{\mathbb{R}}

\newcommand{\norm}[2][]{\|{#2}\|_{#1}}

\newcommand{\eps}{\varepsilon}

\newcommand{\suml}[2]{\sum\nolimits_{#1}^{#2}}

\title{Learning $\ell_1$-based analysis and synthesis sparsity priors using bi-level optimization}

\author{
Yunjin Chen  ~~~Thomas Pock  ~~~Horst Bischof \\
Institute for Computer Graphics and Vision, Graz
    University of Technology \\
Graz A-8010, Austria \\
\texttt{\{cheny, pock, bischof\}@icg.tugraz.at} \\
}

\nipsfinalcopy 

\begin{document}

\maketitle
\begin{abstract}
We consider the analysis operator and synthesis dictionary learning problems based on the 
the $\ell_1$ regularized sparse representation model. We reveal the internal relations between the 
$\ell_1$-based analysis model and synthesis model. We then 
introduce an approach to learn both analysis operator and synthesis dictionary 
simultaneously by using a unified framework of bi-level optimization. 
Our aim is to learn a meaningful operator (dictionary) such 
that the minimum energy solution of the analysis (synthesis)-prior based model is as close as possible to the ground-truth. 
We solve the bi-level optimization problem using the implicit differentiation technique. 
Moreover, we demonstrate the effectiveness of our leaning approach by applying the learned analysis operator (dictionary) to the 
image denoising task and comparing its performance with state-of-the-art methods. 
Under this unified framework, we can compare the performance of the two types of priors.
\end{abstract}

\section{Introduction}
Maximum a Posteriori (MAP) inference under the Bayesian framework 
is a popular method for solving various inverse problems in image processing. 
The MAP estimator is equivalent to an energy minimization problem, which consists of a 
data fidelity term 
and a signal prior term (also known as regularization term). Roughly speaking, the priors fall into two main prior types. 
One is the analysis-based prior and the other is the synthesis-based one.

\textbf{Notation:}
In this paper our model presents a global prior over the entire image, in contrast to the common patch-based one. 
In order to distinguish between 
a patch and an image, we use the notation $\patchx \in \R^m$ to indicate a patch (patch size: $\sqrt{m} \times \sqrt{m}$, $m$ 
is odd), 
and $\imgX \in \R^{MN}$ to indicate an image (image size:$M \times N$, with $m \ll M, m \ll N$). We refer 
$D \in \R^{m\times n}$ and $A \in \R^{n\times m}$ with $m\le n$ to the patch-based synthesis dictionary and analysis operator 
respectively. Furthermore, when the analysis operator $A$ is applied to the entire image $\imgX$, we use the common sliding-window 
fashion to compute the coefficients $Ax$ for all $MN$ patches in the 2-D image form of $u$. This result is equivalent to a multiplication 
of a sparse matrix $\cA \in \R^{(n \times MN) \times MN}$ and $\imgX$, i.e., $\cA \imgX$. 
We can group $\cA$ to $n$ separable sparse matrices $\{\cA_1,\dots,\cA_n\}$, where $\cA_i \in \R^{MN \times MN}$ 
is associated with the $i^{th}$ row 
of $A$ ($A_i$). If we consider $A_i$ as a 2-D filter ($\sqrt{m} \times \sqrt{m}$), we have: $\cA_i u$ is equivalent to the result of 
convolving image $u$ with filter $A_i$. 
Finally, we use $\cA$ that is expanded from the patch-based analysis operator $A$, 
to denote the global analysis operator associated with an entire image. 

\textbf{Patch based analysis and synthesis model:}
Under the framework of MAP, the patch-based analysis model 
is given as the following minimization problem
\begin{equation}\label{analysismodel}
\patchx^* = \arg\min\limits_{\patchx \in \R^m} 
\phi(A\patchx)  + 
\frac{\lambda}{2}\norm[2]{\patchx - f}^2, 
\end{equation}
where $A$ is called analysis operator. The form of the penalty function $\phi$ depends on 
the prior utilized. For sparse representation, it can be $\|\cdot\|_p^p (p \in [0,1])$ or $\log(1+|x|)$. 
The second type of prior is so-called synthesis prior. Basically, in the synthesis-based sparse 
representation model, a signal $\patchx$ is called sparse over a given dictionary $D$, when it can be approximated as a linear 
combination of a few atoms from dictionary $D$. This is formulated 
as following minimization problem using the MAP estimator. 
When we concentrate on the sparse prior, normally the penalty function $\phi$ is chose as $\|\cdot\|_p^p (p \in [0,1])$. 
\begin{equation}\label{synthesismodel}
x = D\alpha^*; \alpha^* = \arg\min\limits_{\alpha \in \R^n} 
\phi(\alpha)  + 
\frac{\lambda}{2}\norm[2]{D\alpha - f}^2. 
\end{equation}
\textbf{Learning patch based analysis and synthesis prior:}
In order to pursue better performance, an intuitive possibility is to make a better choice for the analysis operator $A$ and dictionary $D$ 
based on training. Indeed, there exist several typical and successful training algorithms for over-complete dictionary learning: (i) the 
K-SVD algorithm \cite{KSVDdenoising2006,KSVD2006} (ii) On-line dictionary learning algorithm \cite{MairalBPS09} 
(ii) efficient sparse coding algorithms \cite{Lee2006}. 
However, compared to the extensive study for the training of the synthesis dictionary, the analysis operator learning problem has received 
relatively much less attention in the past decade, although the analysis model is the counterpart to the celebrated synthesis sparse model. 
But fortunately, it has been gaining more and more attention these two years. Consequently, 
there appear different algorithms for analysis operator learning \cite{OphirSequentialLearning,RubinsteinKSVD,
YahoobiAnalysisLearning,YahoobiNoiseAware,YahoobiConstrainedLearning,HaweAnalysisLearning,FadiliAnalysisLearning}.
Among existing analysis operator learning algorithms, the learning approach proposed by Peyr{\'e} and Fadili is very 
appealing since they consider this problem from a novel point of view. They 
interpret the action of analysis operator as convolution with some finite impulse response filters and  
they formulate the analysis operator learning task as a bi-level optimization problem \cite{bileveloverview} 
which is solved using a gradient descent algorithm.

\textbf{Contributions:}
Based on the investigation of existing dictionary and analysis operator learning algorithms, we find that 
(1) all the training approaches are based on patch priors; (2) the study of the later is immature since 
so far only few prior work has been tested with natural images 
\cite{YahoobiNoiseAware,YahoobiConstrainedLearning,HaweAnalysisLearning}; and (3) most analysis 
operator learning algorithms have to impose some 
non-convex constraints on the operator $A$; this therefore makes 
the corresponding optimization problems relatively complex and difficult to solve. 
Thus three questions arise: (1) can we formulate the image-based model using the patch priors? (2) 
is it possible to formulate the analysis operator learning problem in a relatively easy way? 
(3) can we compare two types of priors under an unified framework?
We give answers to these questions in this paper. 
\section{Analysis operator and dictionary learning via bi-level optimization}

\textbf{From patch-based model to image-based one:}
In this paper, we concentrate on convex $\ell_1$ sparse representation. 
In the case of analysis model, following the filter-based MRF model for image restoration, it is straightforward 
to extend the patch-based analysis model to the image-based one, which is given as:
\begin{equation}\label{energyfunction}
u^* = \arg\min\limits_{\imgX}
E(\imgX) = 
\suml{i=1}{n} \|\cA_i\imgX\|_1 + 
\frac{\lambda}{2}\|\imgX-\imgY\|_2^2= 
\|\cA\imgX\|_1 + 
\frac{\lambda}{2}\|\imgX-\imgY\|_2^2, 
\end{equation}
where $\cA$ is the global analysis operator constructed from 
the local patch-based analysis operator $A$, $u$ and $f$ are images ($M \times N$). 
However, if we want to extend the patch-based synthesis model to the image-based one, we find it not as easy as the analysis case. 
Considering the common strategy that averages over-lapping patches, 
we can make explicit use of this strategy of patch-averaging to reconstruct the recovered image, then we arrive at 
our image-based synthesis model
\begin{equation}\label{synthesis}
\{\alpha_{ij}^*\} = \arg\min\limits_{\alpha_{ij}} 
\suml{ij}{}\|\alpha_{ij}\|_1 + 
\frac{\lambda}{2}\|\frac{1}{m}\suml{ij}{}R_{ij}^TD\alpha_{ij} - f\|_2^2, 
\end{equation}
where the size of image $f$ is $M \times N$, the patch size is $\sqrt{m} \times \sqrt{m}$, matrix $R_{ij}$ is an 
$m \times N_p (N_p = M \times N)$
matrix that extracts the $(i,j)$ patch from the image, and $\alpha_{ij}$ is a $n \times 1$ vector.
We explicitly average all the over-lapping patches by a factor $m$, because $\suml{ij}{}R_{ij}^TR_{ij} = m\cI_{N_p \times N_p}$ 
(the number of patches is equal to the number of pixels using symmetrical boundary condition). 
Note that in our formulation, $\alpha_{ij}$ is not independent any more, in contrast to 
their independence in \cite{KSVDdenoising2006}.
If we stack all the $\alpha_{ij}$ and $R_{ij}$ to a huge column vector $\alpha$ and 
a huge matrix $R$ respectively, and construct a huge diagonal-block matrix $\textit{\textbf{D}}$ by using dictionary $D$, \eqref{synthesis} can be 
rewritten as
\begin{equation}\label{synthesis2}
\alpha^* = \arg\min\limits_{\alpha} 
\|\alpha\|_1 +
\frac{\lambda}{2}\|\frac{1}{m}R^T\textit{\textbf{D}}\alpha - f\|_2^2 
= \|\alpha\|_1 +
\frac{\lambda}{2}\|\cD\alpha - f\|_2^2, 
\end{equation}
where $\cD = \frac{1}{m}R^T\textit{\textbf{D}}$. Now we can see the image-based model has the unified form with the 
patch-based one, which 
has a nice MAP interpretation. However, this formulation involves 
too many unknown variables ($n \times N_p$), compared to $N_p$ unknown variables for the analysis model. This is a big drawback 
for our training scheme. we expect to formulate it by $N_p$ variables. Indeed, we succeed by considering its dual problem. 
We introduce a auxiliary variable $u = \cD\alpha$ into the $\ell_2$ norm, and use $v$ to denote the Lagrange multiplier, by using 
definition of the convex conjugate function to the $\ell_1$ norm~\cite{convexBoyd}, we arrive at 
\begin{equation}\label{dualsynthesis2}
v^* = \arg\min\limits_{v} \delta(\cD^Tv) + \frac{1}{2\lambda}\|v - \lambda f\|_2^2,\quad u = f - v/\lambda,
\end{equation}
where the function $\delta(\cD^Tv)$ denotes the indicator function of the interval $[-1,1]$. 
After having a closer look at the connection between the primal variable $\alpha$ and the dual variable $v$, we find that $v$ is exactly 
the additive noise itself, because the recovered image is given by $\cD\alpha^*$ with $\cD\alpha^* = u^* = f - v^*/\lambda$. 
More interestingly, after expanding $\cD^Tv = \textit{\textbf{D}}^T(Rv)/m$, we find that this result is surprisingly equivalent to 
filter response result when applying the analysis operator $\frac{1}{m}D^T$ to an image $v$. Then we draw the conclusion that 
the synthesis dictionary $D$ can also be 
interpreted as an analysis operator $A_D = \frac{1}{m}D^T$. If we use the notation $\cA_D$ to 
denote the global analysis operator as used in 
the aforementioned image-based analysis model, we can present the similarity 
between the $\ell_1$-based analysis and synthesis model, which is given as
\begin{equation}\label{comparison}
\begin{cases}
v^* = \arg\min\limits_{v} \delta(\cA_D v) + \frac{1}{2\lambda}\|v - \lambda f\|_2^2,~ u^* = f - v^*/\lambda, 
~\text{($\ell_1$-based synthesis model)}\\
\imgX^* = \arg\min\limits_{\imgX}
\|\cA\imgX\|_1 ~~~~+ ~~
\frac{\lambda}{2}\|\imgX-\imgY\|_2^2 \hspace{2.65cm}\text{($\ell_1$-based analysis model)}.
\end{cases}
\end{equation}
\textbf{Bi-level framework for synthesis dictionary and analysis operator learning:}
Motivated by the work presented in \cite{FadiliAnalysisLearning}, and the successful training instance to learn optimized parameters of 
MRF model \cite{SamuelFoE}, we propose our analysis operator (dictionary) 
learning approach based on the unified bi-level optimization framework. 
Equation \eqref{comparison} is so-called lower-level problem in our bi-level framework, and we need to define an 
upper-level problem, also known as loss function. Following the work of~\cite{SamuelFoE}, we use the differentiable loss function 
\begin{equation}\label{lossfunction}
L(\imgX^*) = \frac{1}{2}\|\imgX^*-\imgG\|_2^2, 
\end{equation}
where $\imgG$ is the ground-truth image and $\imgX^*$ is the minimizer of energy
function \eqref{comparison}.
Given $S$ training samples $\{\imgY_k,\imgG_k\}_{k=1}^S$, where $\imgG_k$ and $\imgY_k$
are the $k^{th}$ clean image and the corresponding noisy version respectively, our bi-level model aims to learn an meaningful analysis 
operator (dictionary) such that the overall loss function for all samples is as small as possible. 
Therefore, our learning model is formally formulated as the following \textbf{unconstrained} bi-level optimization problem (take analysis operator 
learning model for instance; 
the dictionary learning model is similar).
\begin{equation}\label{learningmodel}
\begin{cases}
\min\limits_{A}L(\imgX^*(A)) = 
\suml{k=1}{S}L_k(\imgX_k^*(A)) = 
\suml{k=1}{S}\frac{1}{2}\|\imgX_k^*(A)-\imgG_k\|_2^2\\
\st \imgX_k^*(A) = \arg\min\limits_{\imgX}E(\imgX,\imgY_k;A) = 
\suml{i=1}{n} \|\cA_i\imgX\|_1 + 
\frac{\lambda}{2}\|\imgX-\imgY_k\|_2^2.
\end{cases}
\end{equation}
\textbf{Advantage of our model:} The most appealing property of our approach is that 
it is not necessary to impose any constraint set over the analysis operator $A$.  
Our training model can avoid trivial solutions naturally, e.g., if $A = 0$, the optimal solution of the energy 
function of \eqref{learningmodel} is certainly $\imgX_k^*(A) = \imgY_k$, which makes the loss 
function still large; thus this trivial solution is not acceptable since the target of our model is 
to minimize the loss function. Therefore, the learned operator $A$
must contain some meaningful filters such that the minimizer of the lower-level problem is close to the ground-truth. 

\textbf{Solving the bi-level problem using implicit differentiation:}
Following the work of \cite{SamuelFoE}, we can compute the gradient of the loss function w.r.t the parameter $A$ 
by using implicit differentiation. 
In order to employ the implicit differentiation rule, we need differentiable penalty functions. We have 
\begin{equation}\label{gx}
\begin{aligned}
\|\cdot\|_{1,\eps}:
\phi(z)~ = \sqrt{z^2 + \eps^2} & \quad
\delta_\eps:
\phi(z) ~= \frac{1}{2\eps}\max(|z|-1,0)^2. 
\end{aligned}
\end{equation}
In our training, we concentrate on mean-zero filters to keep consistent with the findings in the work~\cite{Huang1999_Statistics}; 
therefore, we express the filter $\cA_i$ as a linear combination of a set of basis filters $\{B_1, \dots,B_
{N_B}\}$, i.e., $\cA_i = \suml{j=1}{N_B}\theta_{ij}B_j$. 
Then we obtain the derivatives of the loss function with respect to parameters $\theta_{ij}$, which is given as 
\begin{equation}\label{Xtheta}
\nabla_{\theta_{ij}} L = 
\suml{k=1}{S}\{
-\left(B_j^T \phi'\left(\cA_i \imgX^*\right) + \cA_i^T D_i B_j \imgX^*\right)^T
(\suml{i=1}{n}\cA_i^TD_i\cA_i + \cI)^{-1}
(u^* - g)\}_k,
\end{equation}
where $\phi'(\cA_i\imgX)$ is an $N_p \times 1$ vector obtained by applying function $\phi'(z)$ element-wise to the 
vector $\cA_i\imgX$, and $D_i$ is an 
$N_p \times N_p$ diagonal matrix with each $[D_i]_{n,n}$ entry given by applying the function 
$\phi''(z)$ element-wise to the vector $\cA_i \imgX$. In this formulation, we eliminate the parameter $\lambda$ for simplicity, since 
it can be incorporated into the norm of the analysis operator $A$. 
As given by~\eqref{Xtheta}, we have collected all the necessary information to compute the required gradients, 
then we can employ the gradient descent based algorithms for optimization. In this paper, we make use of an efficient 
quasi-Newton's method, L-BFGS\cite{BFGS}.

\section{Learning experiments and application results for image denoising}
We conducted our training experiments using the training images from the BSDS300\cite{amfm_pami2011} 
image segmentation database. We used the whole 
200 training images, and randomly sampled one $64 \times 64$ patch from each training image, giving us a total of 200 training 
samples. We then generated the noisy versions by adding Gaussian noise with standard deviation $\sigma = 15$. 
In our experiments, we learned an analysis operator $A \in \R^{98 \times 49}$ and synthesis dictionary 
$D \in \R^{49 \times 98}$ from the given training samples. 
In order to guarantee the property of mean-zero, each atom in $A$ or $D$ is expressed as the linear 
combination of the DCT-7 basis excluding the first filter with uniform entries. 

After we learned an meaningful operator $A$ and dictionary $D$, we applied them to the image denoising problem based on the same 
68 test images used in~\cite{RothB09}. 
Tab.~\ref{tab:summary} presents the comparison of the average denoising results achieved by our 
$\ell_1$-based analysis and synthesis model with (i) one state-of-the-art denoising method BM3D \cite{BM3D} 
(ii) the K-SVD approach \cite{KSVDdenoising2006} and (iii) the total variation (TV)-based ROF denoising 
model \cite{pdpock}.
We would like to point out that the TV based approach is the most commonly used $\ell_1$-based analysis operator; 
the K-SVD approach is a synthesis sparse representation model based on $\ell_0$ optimization; 
BM3D is one current state-of-the-art denoising approach which is an image based, not generic prior based method, 
and is a specialized denoising algorithm. 
Fig.~\ref{scatterplots} presents a detailed comparison between our $\ell_1$-based analysis model and 
our $\ell_1$-based synthesis model along with three considered denoising methods over 68 test images for $\sigma = 25$. 
A point above the line means better performance than our $\ell_1$-based analysis model. (Due to space limits, we can not present this 
figure in a large scale. Please refer to the digital version for better visibility.)

\begin{minipage}{18cm}
\begin{minipage}[h]{4cm}
\includegraphics[width=0.7\textwidth]{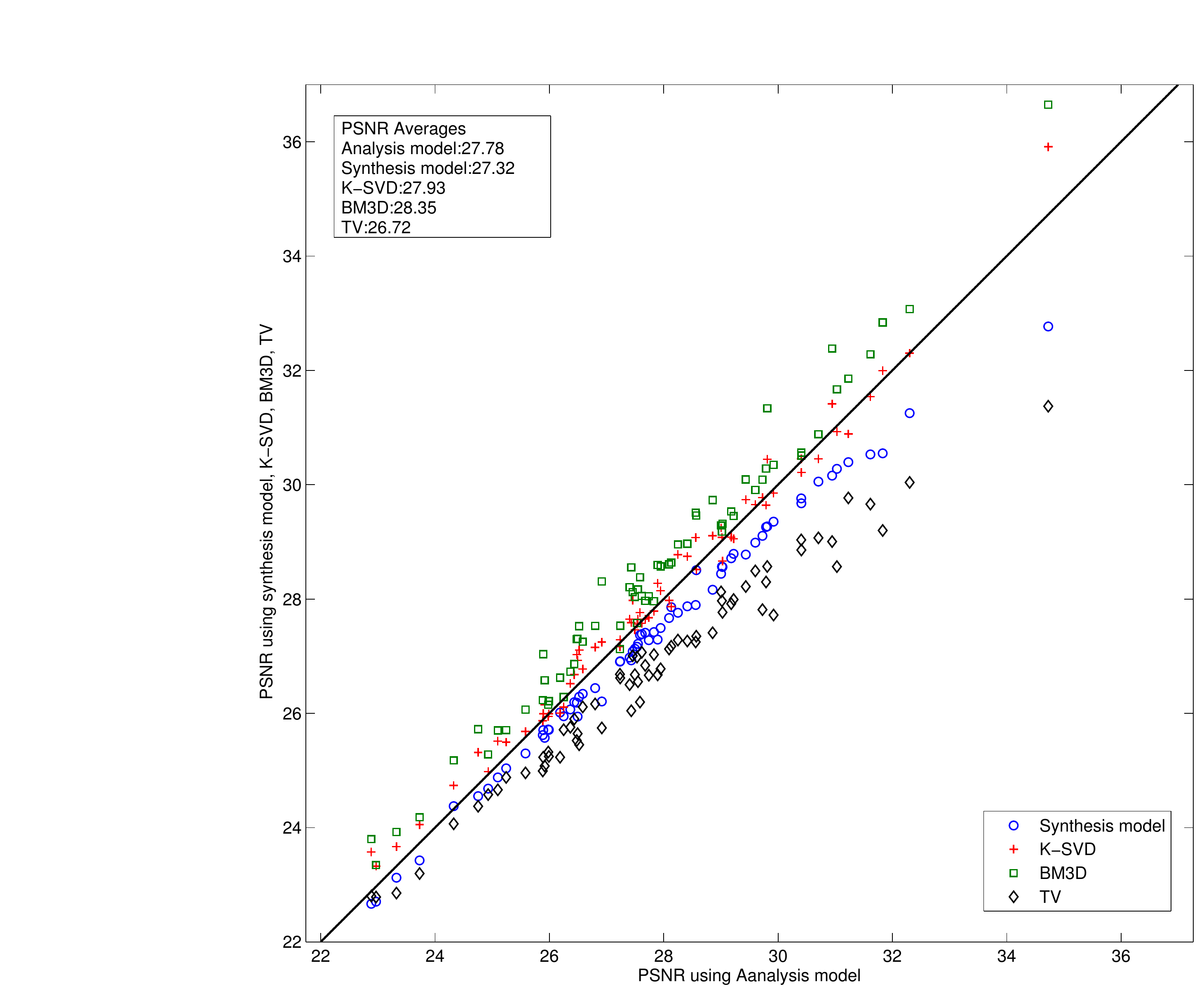}
\makeatletter\def\@captype{figure}\makeatother \caption
{Scatter-plots}
\label{scatterplots}
\end{minipage}
\begin{minipage}[h]{6cm}
\begin{tabular}{|l ||l ||l ||l ||l || l|}
\Xhline{1.2pt}
Model & TV & K-SVD & BM3D & Analysis & Synthesis\\
\Xhline{1.2pt}
avg. PSNR  & 26.72 & 27.93 & 28.35 & 27.78 & 27.32\\
\cline{1-6}
\end{tabular}
\makeatletter\def\@captype{table}\makeatother\caption{Averages of denoising results for 68 test images ($\sigma = 25$)}
\label{tab:summary}
\end{minipage}
\end{minipage}

\section{Conclusions and future work}
From Tab.~\ref{tab:summary} and Fig.~\ref{scatterplots} we can draw 
the following conclusions: (i) the $\ell_1$-based analysis model is significantly superior to the $\ell_1$-based synthesis 
model which is coherent with the findings in the work \cite{EladAnalysisVSSynthesis}. We believe the essential reason lies in the ineffectual 
way the $\ell_1$-based synthesis model characterizes the natural images, since it tries to model the noise signal, not the natural image itself as  
aforementioned. This inferiority also appeared in the training. (ii) our analysis model is comparable with the $\ell_0$-based synthesis model 
K-SVD, as can be seen in Fig.~\ref{scatterplots}.
Compared with specialized methods for image denoising task such as BM3D, our $\ell_1$-based analysis model still can not compete. 
However, its denoising performance is always 
significantly better than the TV based approach.

It is well known that the probability density function (PDF) of the response of zero mean linear filters 
on natural images has heavily tailed distribution \cite{Huang1999_Statistics}. 
Therefore, our future work will concentrate on non-convex penalty function such as $\sqrt{|z|}$ or $\log(1+|z|)$. 
\textbf{According to our preliminary experience about the analysis model using $\log(1+|z|)$ as penalty function, it clearly outperforms the 
$\ell_0$-based synthesis model K-SVD, and has already been on par with BM3D. (We will present this result in our future work.)} 
However, for the case of non-convex, since the Fenchel's duality we used 
in this paper is not available any more, how to handle the synthesis model becomes a problem. This will be the subject of our future work.

\bibliographystyle{plain}
\bibliography{bilevel_learning}
\end{document}